**Title:** Image reconstruction by domain transform manifold learning

**Authors:** Bo Zhu[1,2,3], Jeremiah Z. Liu[4], Bruce R. Rosen[1,2], Matthew S. Rosen[1,2,3*]

[1]A.A. Martinos Center for Biomedical Imaging, Massachusetts General Hospital, Boston, MA, [2]Harvard Medical School, Boston, MA, [3]Dept. of Physics, Harvard University, Cambridge, MA, [4]Department of Biostatistics, Harvard University, Cambridge, MA

* to whom correspondence should be addressed

**Image reconstruction plays a critical role in the implementation of all contemporary imaging modalities across the physical and life sciences including optical[1], radar[2], magnetic resonance imaging (MRI)[3], X-ray computed tomography (CT)[4], positron emission tomography (PET)[5], ultrasound[6], and radio astronomy[7]. During an image acquisition, the sensor encodes an intermediate representation of an object in the sensor domain, which is subsequently reconstructed into an image by an inversion of the encoding function. Image reconstruction is challenging because analytic knowledge of the exact inverse transform may not exist *a priori,* especially in the presence of sensor non-idealities and noise. Thus, the standard reconstruction approach involves approximating the inverse function with multiple *ad hoc* stages in a signal processing chain whose composition depends on the details of each acquisition strategy, and often requires expert parameter tuning to optimize reconstruction performance. We present here a unified framework for image reconstruction, AUtomated TransfOrm by Manifold APproximation (AUTOMAP), which recasts image reconstruction as a data-driven, supervised learning task that allows a mapping between sensor and image domain to emerge from an appropriate corpus of training data. We implement AUTOMAP with a deep neural network and exhibit its flexibility in learning reconstruction transforms for a variety of MRI acquisition strategies, using the same network architecture and hyperparameters. We further demonstrate its efficiency in sparsely representing transforms along low-dimensional manifolds, resulting in superior immunity to noise and reconstruction artifacts compared with conventional handcrafted reconstruction methods. In addition to improving the reconstruction performance of existing acquisition methodologies, we anticipate accelerating the discovery of new acquisition strategies across imaging modalities as the burden of reconstruction becomes lifted by AUTOMAP and learned-reconstruction approaches.**

The paradigm shift from manual to automatic feature extraction in a host of machine learning tasks including speech recognition[8] and image classification[9] has demonstrated the power of allowing real-world data to guide efficient representation through a structured training process. This strategy is mirrored in biological organisms for refining visual perception in a process known as perceptual learning[10]. Human visual reconstruction of time-domain neural codes into the percept image is trained through experience during cognitive development into adulthood, and this conditioning on prior data has been shown to be critical in its robust performance in difficult settings such as low signal-to-noise (SNR)[11] — a fundamentally challenging regime for *artificial* imaging systems across modalities. For example, in contemporary medical imaging, faithful reconstruction of noisy image acquisitions is of particular importance as the clinical push for faster scanning increasingly relies upon reduced-SNR acquisition techniques, be they undersampled MRI, or low dose CT imaging. This problem is compounded by the variety of disparate image encoding strategies in use, each of which require corresponding sophisticated reconstruction chains that are custom-built to craft an approximate inversion of the encoding function (Fig. 1a). Even generalized reconstruction formalisms (e.g., SPIRiT[12] and CG-SENSE[13] for MRI) assume specific encoding spaces (i.e. Fourier), and typically rely upon iterative optimization methods that can amplify artifacts unaccounted for in explicitly imposed noise models.

Inspired by the perceptual learning archetype, we have developed a data-driven unified image reconstruction approach, **Au**tomated **T**ransf**o**rm by **M**anifold **Ap**proximation (AUTOMAP), that learns a near-optimal reconstruction mapping between the sensor domain data and image domain output (Fig. 1a). As this mapping is trained, a low-dimensional joint manifold of the data in both domains is implicitly learned (Fig. 1b), capturing a highly expressive representation that is robust to noise and other input perturbations.

We implemented our unified reconstruction framework with a deep neural network feed-forward architecture composed of fully-connected layers followed by a sparse convolutional autoencoder (Fig. 1c). The fully-connected layers approximate the between-manifold projection from the sensor domain to the image domain. The convolutional layers extract high-level features from the data and force the image to be represented sparsely in the convolutional feature space. Our network operates similarly to the denoising autoencoder described in Vincent et al.[14], however rather than finding an efficient representation of the identity to map $f(x) = \phi_x \circ \phi_x^{-1}(x) = x$ over the manifold of inputs $\mathcal{X}$ (where $\phi_x$ maps the intrinsic coordinate system of $\mathcal{X}$ to Euclidean space near $x$), AUTOMAP extends this framework by finding both a between-manifold projection $g$ from $\mathcal{X}$ (the manifold of sensor inputs) to $\mathcal{Y}$ (the manifold of output images), and a manifold mapping $\phi_y$ to "decompress" the image from manifold $\mathcal{Y}$ back to the representation in Euclidean space. A composite transformation $f(x) = \phi_y \circ g \circ \phi_x^{-1}(x)$ over the joint manifold $\mathcal{M}_{\mathcal{X},\mathcal{Y}} = \mathcal{X} \times \mathcal{Y}$ (Fig. 1b) is therefore achieved. A full mathematical description of this manifold learning process is detailed in the supplemental text.

We demonstrate the AUTOMAP approach using MRI as a model system, though we emphasize that AUTOMAP is generally applicable to image reconstruction problems across a broad range of modalities. The plethora of MRI acquisition strategies makes it a particularly appropriate platform to exhibit the flexibility of AUTOMAP reconstruction over a variety of encoding schemes. We first evaluated the performance of AUTOMAP alongside conventional methods in four nontrivial reconstruction tasks: 1) Radon projection imaging (the standard encoding for CT and PET) and the Kaczmarz-iterative Algebraic Reconstruction Technique[15]; 2) Spiral-trajectory *k*-space (rapid acquisition with non-Cartesian sampling) and Conjugate-Gradient SENSE reconstruction employing NUFFT regridding[13]; 3) Poisson-disc undersampled *k*-space (incoherent sparse acquisition) and compressed sensing reconstruction with wavelet sparsifying transform[16]; 4) Misaligned *k*-space (commonplace sampling distortion due to miscalibrated hardware) and the conventional Inverse Fast Fourier Transform. Evaluation of the AUTOMAP network was performed on brain MR images selected from the Human Connectome Project (HCP)[17], which were transformed to the sensor domain according to the four encoding schemes (see Supplementary Methods for data preparation details) and with varying levels of additive white Gaussian noise introduced to observe reconstruction performance in noisy conditions.

All reconstruction tasks employed the same network architecture and hyperparameters – only the input and output training data differed. To demonstrate AUTOMAP's remarkable generalization capability, all reconstruction tasks except the undersampled encoding were trained from datasets derived entirely from photographs of natural scenes from ImageNet[18] as schematically portrayed in Fig. 1b; for these acquisitions, the network was not exposed to any MRI or other medical images until the test phase (see Supplementary Methods for data preparation and training details).

The results shown in Fig. 2 demonstrate the ability of AUTOMAP to faithfully reconstruct across varying encoding acquisition strategies. We emphasize here that the reconstruction transforms emerged strictly from training on data samples, without any higher-level concepts (e.g. mathematical transforms or domain representations) introduced at any stage. To learn a new reconstruction for a particular encoding acquisition, one simply needs to generate a training dataset with the encoding forward model. The ability for AUTOMAP to represent a variety of sophisticated transform functions with a single network architecture is grounded in the inherent universal approximation properties of nonlinear multilayer perceptron systems[19,20].

Furthermore, AUTOMAP outperforms conventional methods, in particular exhibiting superior noise and sampling-defect immunity as shown in Fig. 2. Artifacts appear only in the conventional reconstructions: white noise amplification for iterative inverse-Radon[21], ringing artifacts due to NUFFT regridding of noisy samples[22], complex artifacts from noisy undersampled compressed sensing reconstruction[23], and substantial aliasing from misaligned sampling trajectories[24]. Additive Gaussian noise was not injected during training; the noise-immunity we observe was not trained explicitly, or imposed by predictive noise-modeling, but rather inherently resulted from the manifold learning process extracting robust features of the data, leading to improvement in SNR during reconstruction. This emphasis on modeling features of the signal rather than noise statistics for low-SNR performance is consistent with the neural mechanisms underlying human visual perceptual learning[25].

We next examined the hidden-layer activity of our AUTOMAP network during the feed-forward reconstruction process. We trained AUTOMAP using training data derived from either ImageNet, HCP brain images, or random-valued Gaussian noise without any real-world image structure. Each trained network was then used to reconstruct the fully-sampled Cartesian *k*-space of a single brain image. The activation values of the hidden-layer FC2 (Fig. 1c) are plotted in (Fig. 3a-c). As the training moves from general (Fig 3a) to specific (Fig 3c), we observe the hidden-layer activity exhibiting greater sparsity, indicating successful extraction of robust features[26], consistent with the noise immunity observed in our experiments. We note that fully-connected hidden-layer sparsity was not explicitly imposed (*i.e.* not enforced by a penalty in the loss function), but emerged naturally through the training process. A normalized histogram of the hidden-layer activations is shown in Fig. 3d. A representative set of the convolutional kernels applied to feature maps in layer C2 (Fig. 1c) is shown in Fig. 3h. Processing by the convolutional layers is similar to that of compressed sensing, however instead of assuming an explicit sparsifying transform (e.g., wavelet), AUTOMAP simultaneously learns an optimal convolutional domain and the sparse representations through a joint optimization (see Supplementary Methods for details).

We then studied the weight parameters of each trained network using a t-SNE analysis[27] (Fig 3e-g) which embeds a high-dimensional dataset into a low-dimensional space for visualization. Here we visualize the spatial relationship of trained network weights, particularly from FC2 to the pre-convolutional FC3 image layer (see Supplementary Methods for details). Fig. 3d shows that the t-SNE embedding of the noise-trained network weights is highly unorganized and unbiased with respect to pixel location; this is unsurprising because the network learns a "pure" or "neutral" Fourier Transform that does not recognize the local spatial correlation that exists in real-world images[28]. The generic-image-trained model respects this local spatial correlation, thus the weights to neighboring pixels are more similar, as shown by the t-SNE embedding (Fig. 3f). This feature is most clearly exhibited by the t-SNE of the brain-trained network weights (Fig. 3g)., which are organized into a two-dimensional sheet (in the three-dimensional embedding space), demonstrating extremely high similarity between weights to two-dimensional neighbors for all pixel locations.

Lastly, we demonstrated AUTOMAP's ability to learn reconstruction of image phase from complex-valued sensor data by including phase-modulated data in the network training. We created a phase-modulated training set by generating synthetic phase patterns (examples shown in Fig. 4a) to modulate the magnitude-only training images collected from the HCP database prior to encoded in *k*-space. Using the same *k*-space data as input, we trained separate AUTOMAP networks to reconstruct magnitude and phase with their respective target training images, and validated this on reconstructing *in vivo k*-space (Fig. 4d-e). This reconstruction can also be performed on one larger network with concatenated magnitude and phase data. This phase-modulation training allows public medical image databases and private PACS repositories to be used for training, despite the typical absence of phase data in these datasets. Furthermore, our results show that parametric influences on the input signal can be simulated for training and subsequently disentangled by AUTOMAP, suggesting utility towards sophisticated reconstruction problems such as automated motion compensation.

AUTOMAP provides a powerful new paradigm for image reconstruction implemented with a deep neural network that learns an optimal reconstruction function for any acquisition strategy. By conditioning this reconstruction upon low-dimensional manifolds defined by real-world data, AUTOMAP exhibits superior performance for noisy acquisitions compared with conventional approaches. We anticipate that the noise robustness attainable with our approach will improve imaging quality and speed for a broad range of applications exhibiting low-SNR including low-dose CT[29], low-light charge-coupled devices (CCD)[30], large-baseline radio astronomy[31], and rapid volumetric optical coherence tomography (OCT).[32] Finally, as the reconstruction problem can be solved by AUTOMAP for arbitrary encoding schemes, we also anticipate this enabling the development of new classes of acquisition strategies across imaging modalities.

**Acknowledgements.** We gratefully acknowledge Dr. Mark Michalski and the computational resources and assistance provided by the Massachusetts General Hospital (MGH) and the Brigham and Women's Hospital (BWH) Center for Clinical Data Science (CCDS). The CCDS is supported by MGH, BWH, the MGH Department of Radiology, the BWH Department of Radiology, and through industry partnership with NVIDIA. We also thank David E J Waddington and Ronald L Walsworth for their insightful comments on this manuscript. B.Z. was supported by the National Institutes of Health / National Institute of Biomedical Imaging and Bioengineering F32 Fellowship (EB022390). Data were provided in part by the Human Connectome Project, MGH-USC Consortium (Principal Investigators: Bruce R. Rosen, Arthur W. Toga and Van Wedeen; U01MH093765) funded by the NIH Blueprint Initiative for Neuroscience Research grant; the National Institutes of Health grant P41EB015896; and the Instrumentation Grants S10RR023043, 1S10RR023401, 1S10RR019307.

**Author Contributions.** B.Z., J.Z.L., B.R.R., and M.S.R. conceptualized the problem and contributed to experimental design. B.Z. developed, implemented, and tested the technical framework. J.Z.L. and B.Z. constructed the theoretical description. B.Z., J.Z.L., B.R.R., and M.S.R. wrote the manuscript.

## Methods

**Data acquisition and pre-processing.** The training dataset of generic images was assembled from ImageNet[18]. 10,000 images from the "Animal," "Plant," and "Scene" categories were each cropped to the central 256 × 256 pixels and subsequently subsampled to 128 × 128. Y-channel luminance was extracted from the RGB color images to form grayscale intensity images. Each image was then rotated in 90° increments to augment the dataset. The mean intensity of each image was subtracted and the entire dataset was normalized to a constant value defined by the maximum intensity of the dataset.

The training dataset of brain images was assembled from the MGH-USC Human Connectome Project (HCP)[17] public database, which were acquired with T1-weighted 3D MPRAGE with TR=2530 ms, TE=1.15 ms, TI=1100 ms, FA=7.0°, BW=651 Hz/Px on a Siemens Skyra 3T MRI platform (Siemens Medical Solutions, Erlangen, Germany). Axial, sagittal, and coronal T1-weighted slices from 131 subjects were used to generate a 50,000 image dataset. For each image, the central 255 × 256 pixels were cropped and subsampled to 128 × 128. To promote translation invariance in the training, each image was symmetrically tiled to create a larger 256 × 256 image containing four reflections of the original, and cropped to a random 128 × 128 section. The same data normalization process described above for the ImageNet dataset was used. The test data used in the first evaluation experiment were taken from another subject outside the subjects used for training. The test data also included T2-weighted SPACE acquisitions with TR=3200 ms, TE=561 ms, FOV=224 × 244 mm, BW=744 Hz/Px on the same Siemens Skyra 3T MRI platform.

The *in vivo k*-space test data used in the final experiment was acquired on a 3T Siemens Trio scanner with a spin-echo imaging sequence with TR = 3110 ms, TE=23.0 ms, matrix size = 208 × 256, and slice thickness = 3 mm. Data from the 12-channel receiver head coil was coil-compressed with singular value decomposition (SVD) and the central 128 × 128 *k*-space samples formed the input for the 128 × 128 reconstruction task.

**Sensor-domain encoding.** Sensor domain representations for each image were encoded according to the reconstruction task. For the Radon transform experiment (Fig. 2a-c), we used the Discrete Radon Transform with 180 projection angles and 185 parallel rays; the Spiral *k*-space experiment (Fig. 2d-f) used Nonuniform Fast Fourier Transform (NUFFT)[3] to encode a 10-interleave spiral trajectory[33] with variable density factor $\alpha$=1 and undersampling factor R = 1/1.2 based on the pre-subsampled 256 × 256 images (The MATLAB code used for this trajectory encoding is available at http://bigwww.epfl.ch/algorithms/mri-reconstruction/); the undersampled Cartesian *k*-space experiment (Fig. 2g-i) used a Poisson-disc sampling pattern with 40% undersampling of the Fourier Transformed *k*-space generated with the Berkeley Advanced Reconstruction Toolbox (BART)[34]; the misaligned Cartesian *k*-space experiment (Fig. 2j-l) used Fourier Transformed *k*-space with each readout-direction line randomly shifted in either direction with up to 3 samples of displacement.

**Modulation with synthesized phase.** Phase-modulated data was used in the final reconstruction experiment to train the network to reconstruct phase images. Synthesized phase maps were created by generating two-dimensional sinusoids with varying spatial frequencies independent along each image axis, and rotated by a random angle with respect to the image axes. The intensities of the sinusoids represented phase values, and were normalized to be between 0 and $2\pi$. Each magnitude image in the training dataset was then modulated with a randomly generated phase map to form the complex-valued target image, which was then encoded to *k*-space with FFT to form the sensor domain input.

**Model architecture.** The input to the neural network consists of a vector of sensor domain sampled data produced by the preprocessing steps detailed above. Because the input layer is fully connected to the first hidden layer, for each reconstruction task the sensor domain data (typically represented in two dimensions for images) can be vectorized in any order without any effect on the training. Since the neural network computational framework used here (Tensorflow[35]) operates on real-valued inputs and parameters, complex data must be separated into real and imaginary components concatenated in the input vector. Thus, an n × n complex-valued *k*-space matrix, for example, is reshaped to a $2n^2$ × 1 real-valued vector (for our experiments, n=128). As schematically illustrated in Fig. 1c, the input layer FC1 is fully connected to an $n^2$ × 1 dimensionality hidden layer FC2 and activated by the hyperbolic tangent function. This first hidden layer is fully connected to another $n^2$ × 1 dimensionality hidden layer FC3 with hyperbolic tangent activation, and is reshaped to an n × n matrix in preparation for convolutional processing. The first convolutional layer C1 convolves 64 filters of 5 × 5 with stride 1 followed by a rectifier nonlinearity[36]. The second convolutional layer C2 again convolves 64 filters of 5 × 5 with stride 1 followed by a rectifier nonlinearity. The the final output layer deconvolves the C2 layer with 64 filters of 7 × 7 with stride 1. The output layer represents the reconstructed magnitude image, except for the phase-modulation experiment, where the network was trained separately to reconstruct the real and imaginary components of the image.

**Training details.** The same network architecture and hyperparameters were used for our experiments. For each sensor encoding reconstruction task, a different network was trained from the corresponding sensor domain encodings and target images applied to the inputs and outputs, respectively, of the neural network (details of training data and network architecture described above). Multiplicative noise at 1% was applied the input to promote manifold learning during training by forcing the network to learn robust representations from corrupted inputs[14]. We note that the specific noise distribution of this corruption process did not serve to model the additive Gaussian noise that was applied during evaluation. The RMSProp algorithm (see http://www.cs.toronto.edu/~tijmen/csc321/slides/lecture_slides_lec6.pdf) was used with minibatches of size 100, learning rate 0.00002, momentum 0.0, and decay 0.9. The loss function minimized during training was a simple squared loss between the network output and target image intensity values, with an additional L1 norm penalty ($\lambda$=0.0001) applied to the feature map activations in the final hidden layer C2 to promote sparse convolutional representations. The convolutional layers are inspired by

Winner-Take-All autoencoders[37] which jointly optimize the sparse convolutional codes as well as the deconvolutional kernel "dictionaries" upon which the final image is built (Fig. 3h). Note that this imposed sparsity on the convolutional layers is separate from the fully-connected hidden-layer activation sparsity that emerged without an applied sparsifying penalty (Fig. 3a-c), and occurs even without imposed convolutional sparsity. Each network was trained for 100 epochs on the Tensorflow[35] deep learning framework using 2 NVIDIA Tesla P100 GPUs with 16GB memory capacity each, specifically employing either a conventional server platform with 2 P100 GPUs or the NVIDIA DGX-1 utilizing two GPUs per experiment.

**Evaluation procedure.** The performance of AUTOMAP-trained networks for the four acquisition strategies was evaluated by reconstructing the four sensor domain encodings of T1- and T2-weighted MRI brain images of a human subject from the HCP database as described above. For the Radon transform, Spiral $k$-space, and misaligned $k$-space experiments, the network was trained using ImageNet data; for the undersampled $k$-space experiment, the network was trained with data from the HCP brain image dataset using only T1-weighted images from other subjects in the HCP database.

We reconstructed the same set of sensor domain inputs with conventional reconstruction techniques for each acquisition strategy: For Radon projection imaging, the Algebraic Reconstruction Technique[15,38] with the Kaczmarz method (10 iterations) was used, implemented with MATLAB code available at http://www.imm.dtu.dk/~pcha/AIRtools/; Spiral-trajectory $k$-space was reconstructed with a single-coil implementation of CG-SENSE using NUFFT regridding[13] over 30 conjugate gradient iterations, with MATLAB code available at http://bigwww.epfl.ch/algorithms/mri-reconstruction/; Poisson-disc undersampled $k$-space was reconstructed with compressed sensing[16] using the wavelet sparsifying transform with the L1 penalty parameter $\lambda$=0.01, using the Berkeley Advanced Reconstruction Toolbox[34] with code available at https://mrirecon.github.io/bart/; misaligned $k$-space (commonplace sampling distortion due to miscalibrated hardware) was reconstructed with the native MATLAB implementation of the two-dimensional Inverse Fast Fourier Transform.

To probe the noise sensitivity of the reconstructions, varying levels of additive white Gaussian noise were introduced to the sensor domain encodings: 25 dB SNR for the Spiral experiment, 30 dB SNR for the undersampled Cartesian experiment, and 40 dB SNR for the Radon projection experiment. We did not noise-corrupt the misaligned $k$-space because the sampling trajectory already represented a perturbed input. The phase-modulated reconstruction experiment used *in vivo* Cartesian $k$-space input without additive noise.

**t-SNE analysis.** Relationship of trained network weights were visualized with t-SNE (t-distributed stochastic neighbor embedding)[27]. We employed a standard Cartesian Fourier $k$-space encoding for the networks. To reduce computational load, lower-resolution reconstruction networks were trained using 64 × 64 images from either ImageNet, brain images, or random-valued noise without any real-world image structure. In the visualization, each point corresponds to a single pixel in FC3, represented by an $n^2$- dimensional vector of weights directed to it from the FC2 layer. The label for each point is a scalar pixel location in the image space (from 0 to $n^2$) which also defines its color in the visualization; similar colors correspond to similar pixel location. The t-SNE algorithm was implemented with perplexity 64 over 200 iterations with MATLAB code available at https://lvdmaaten.github.io/tsne/.

**Data and code availability.** The generic natural images used for training are available from the ImageNet database (http://www.image-net.org/). The brain images used for training and evaluation were obtained from the MGH-USC Human Connectome Project (HCP) database (https://db.humanconnectome.org/). Source code is available from the corresponding author upon reasonable request.

**Description of AUTOMAP manifold learning.** Our learning task is two-fold. Given $\tilde{x}$ the noisy observation of sensor domain data $x$, we want to first learn the stochastic projection operator onto $\mathcal{X}$: $p(\tilde{x}) = P(x|\tilde{x})$. After obtaining $x$, our second and more important task is to reconstruct $f(x)$ by producing a reconstruction mapping $\hat{f}: \mathbb{R}^{n^2} \to \mathbb{R}^{n^2}$ that minimizes the reconstruction error $L(\hat{f}(x), f(x))$.

We first describe the reconstruction process by consider the idealized scenario where the input sensor data are noiseless. Denote the data as $(y_i, x_i)_{i=1}^n$, where for $i^{th}$ observation $x_i$ indicates a $n \times n$ set of input parameters, and $y_i$ indicates the $n \times n$ real, underlying images. We assume that

(1) there exists a unknown smooth and homeomorphic function $f: \mathbb{R}^{n^2} \to \mathbb{R}^{n^2}$, such that $y = f(x)$, and
(2) $(x_i)_{i=1}^n$, $(y_i)_{i=1}^n$ lie on unknown smooth manifolds $\mathcal{X}$ and $\mathcal{Y}$, respectively. Both manifolds are embedded in the ambient space $\mathbb{R}^{n^2}$, such that $\dim(\mathcal{X}) < n^2$ and $\dim(\mathcal{Y}) < n^2$.

Above two assumptions combines to define a joint manifold $\mathcal{M}_{\mathcal{X},\mathcal{Y}} = \mathcal{X} \times \mathcal{Y}$ that the entire dataset $(x_i, y_i)_{i=1}^n$ lies in, which can be written as:

$$\mathcal{M}_{\mathcal{X},\mathcal{Y}} = \{(x, f(x)) \in \mathbb{R}^{n^2} \times \mathbb{R}^{n^2} | x \in \mathcal{X}, f(x) \in \mathcal{Y}\}.$$

Note $(x, f(x))$ is described using the regular Euclidean coordinate system, we may equivalently describe this point using the intrinsic coordinate system of $\mathcal{M}_{\mathcal{X},\mathcal{Y}}$ as $(z, g(z))$ such that there exists a homeomorphic mapping $\phi = (\phi_x, \phi_y)$ between $(x, f(x))$ and $(z, g(z))$. (i.e. $x = \phi_x(z)$ and $f(x) = \phi_y \circ g(z)$). As a side note, in topology, $\phi = (\phi_x, \phi_y): \mathcal{M}_{\mathcal{X},\mathcal{Y}} \to \mathbb{R}^{n^2} \times \mathbb{R}^{n^2}$ corresponds to the local coordinate chart of $\mathcal{M}_{\mathcal{X},\mathcal{Y}}$ at the neighborhood of $(x, f(x))$. Instead of directly learning $f$ in the ambient space, we wish to learn the diffeomorphism $g$ between $\mathcal{X}$ and $\mathcal{Y}$ in order to take advantage of the low-dimensional nature of embedded space. Consequently, the process of generating $y = f(x)$ from $x$ can be written as a sequence of function evaluations:

$$f(x) = \phi_y \circ g \circ \phi_x^{-1}(x).$$

For the convenience of later presentation, we notice that given input image $x$, the output image follows a probability distribution $Q(Y|X = x, f)$, which is a degenerate distribution with point mass at $y = f(x)$.

We now turn to the more realistic scenario where corruption exists in the sensor domain input and describe the denoising process. Instead of observing the perfect input data $x_i$, we observe $\tilde{x}_i$ which is a corrupted version of $x_i$ by some known corruption process described by the probability distribution $P(\tilde{X}|X = x)$. In order to handle this complication, we seek to learn a denoising step $Q(X|\tilde{X} = \tilde{x}, p)$ to our model pipeline, such that our prediction for $y$ is no longer a deterministic value, but a random variable with conditional distribution $P(Y|\tilde{X})$ so that we can properly characterize the prediction uncertainty caused by the corruption process.

Instead of learning this denoising step explicitly, we draw analogy from denoising autoencoders[39] and model the joint distribution $P(Y, X, \tilde{X})$ instead. Specifically, in addition to the assumptions (1)-(2) listed above, we also assume:

(3) the true distribution $P(X|\tilde{X})$ lies in the semiparametric family $\mathbb{Q}$ defined by its first moment $\mathbb{Q} = \{Q(X|\tilde{X} = \tilde{x}, p) | E(X) = p(\tilde{X})\}$.

We model $P(Y, X, \tilde{X})$ using the decomposition below:

$$Q_{(f,p)}(Y, X, \tilde{X}) = Q(Y|X, f) Q(X|\tilde{X}, p) P(\tilde{X}),$$

where $Q(Y|X, f)$ denote the model for reconstruction process that we have described earlier, $Q(X|\tilde{X}, p)$ the denoising operator that we seek to learn, and $P(\tilde{X})$ the empirical distribution of corrupted images. Notice that we can combine the models for denoising and reconstruction processes together by collapsing the first two terms on the right-hand side into one term, which gives:

$$Q_{(f,p)}(Y, X, \tilde{X}) = Q(Y, X|\tilde{X}, (f, p)) P(\tilde{X}).$$

We remind readers that $Y = f(X)$ is a deterministic and homeomorphic mapping of $X$, therefore $Q(Y, X|\tilde{X}, (f, p)) = Q(Y|\tilde{X}, (f, p))$ is the predictive distribution of output image $y$ given the noisy input $\tilde{x}$, which is exactly our estimator of interest. Consequently, the model can be written as:

$$Q_{(f,p)}(Y, X, \tilde{X}) = Q(Y|\tilde{X}, (f, p)) P(\tilde{X}).$$

This completes the definition of our model for the joint distribution.

In the actual training stage, we usually took advantage of the fact that perfect input images $x$ are available, and train the model with $\tilde{x}$ that we generated from $P(\tilde{X}|X = x)$. That is to say, the joint distribution of $(Y, X, \tilde{X})$ observed in training data admits the form:

$$P(Y, X, \tilde{X}) = P(Y|X) P(\tilde{X}|X) P(X).$$

The training proceeds by minimizing the KL-divergence between observed probability $P(Y,X,\tilde{X})$ and our model $Q(Y,X,\tilde{X})$,

$$\mathbb{D}_{KL}\{P(Y,X,\tilde{X})||Q_{(f,p)}(Y,X,\tilde{X})\},$$

with respect to the function-valued parameters $(f,p)$. As the KL-divergence converges toward 0, $Q(X|\tilde{X},p)$ converges to $P(X|\tilde{X})$ the denoising projection, and at the same time $Q(Y|\tilde{X},(f,p))$ converges to $P(Y|\tilde{X})$.

There exists a rich literature in explicitly learning of the stochastic projection $p$, diffeomorphism $g$, and the local coordinate chart $\phi$ [40-44]. However, we notice that since $(\phi_f, \phi_x, p, g) \in \mathbb{C}^\infty$ (where $\mathbb{C}^\infty$ denote the set of infinitely differentiable functions), $\hat{f} = \phi_f \circ g \circ \phi_x^{-1} \circ p$ as a whole is a continuously differentiable function on a compact subset of $\mathbb{R}^{n^2}$, and can therefore be approximated with theoretical guarantee by the universal approximation theorem[45].

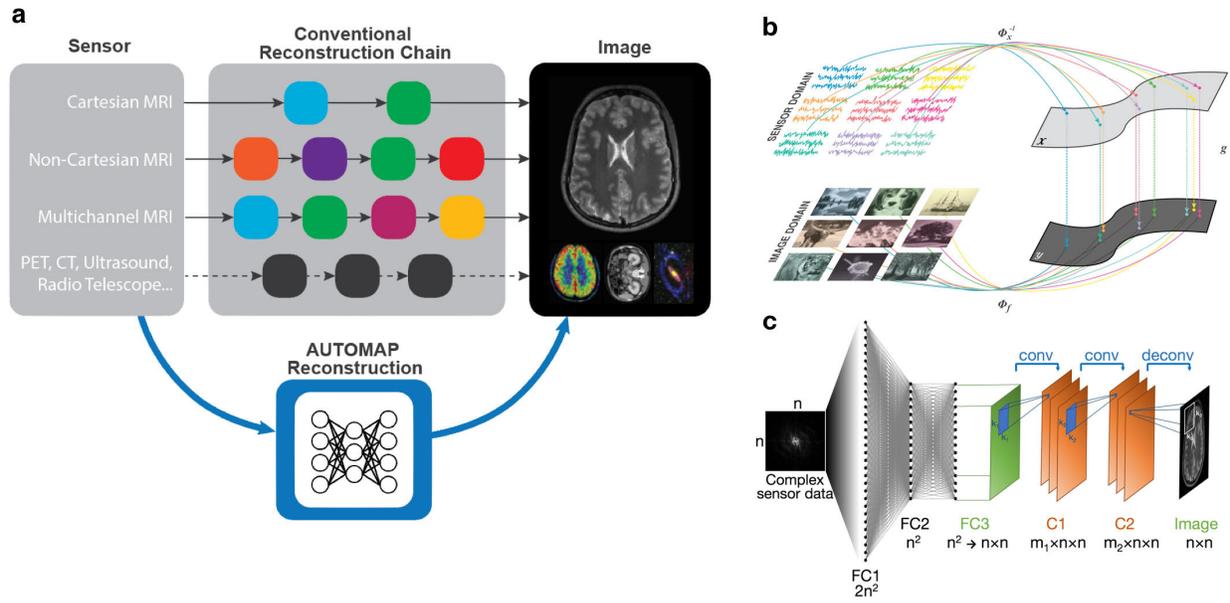

**Figure 1. Schematic representations of AUTOMAP image reconstruction. a,** Conventional image reconstruction is implemented with sequential modular reconstruction chains composed of handcrafted signal processing stages that may include discrete transforms (e.g., Fourier, Hilbert, Radon), data interpolation techniques, nonlinear optimization, and various filtering mechanisms. AUTOMAP replaces this approach with a unified image reconstruction framework that learns the reconstruction relationship between sensor and image domain without expert knowledge. **b,** An optimal mapping between sensor domain and image domain is determined via supervised learning of sensor (top) and image (bottom) domain pairs. The training process implicitly learns a robust low-dimensional joint manifold $\mathcal{X} \times \mathcal{Y}$ over which the reconstruction function $f(x) = \phi_y \circ g \circ \phi_x^{-1}(x)$ is conditioned. **c,** AUTOMAP is implemented with a deep neural network architecture composed of fully-connected layers (FC1 to FC3) with hyperbolic tangent activations followed by a convolutional autoencoder (FC3 to Image) with rectifier nonlinearity activations (see Supplementary Methods for model architecture details).

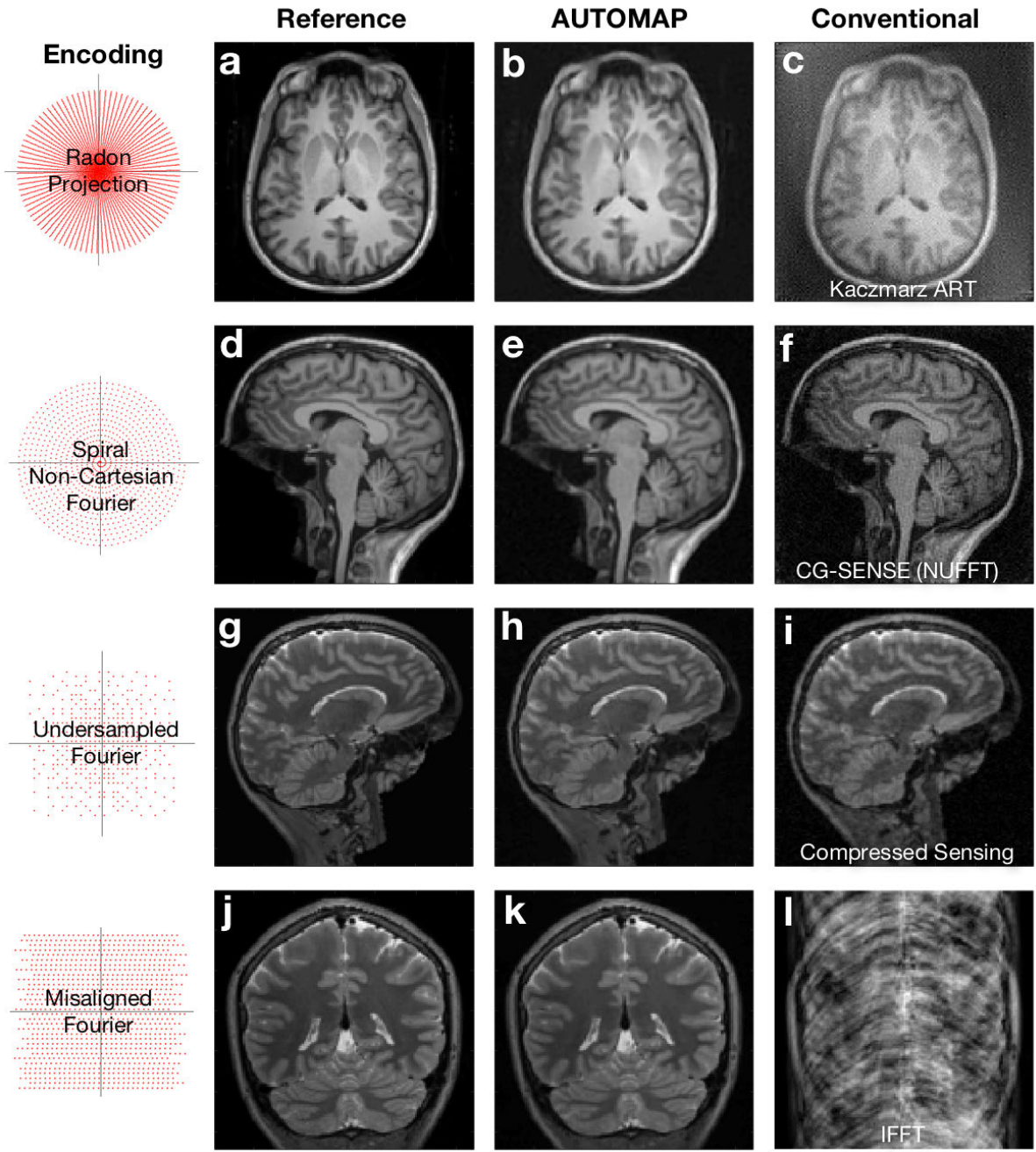

**Figure 2. Reconstruction performance of AUTOMAP compared with conventional techniques.** Reference brain images were encoded into four sensor domain sampling strategies with varying levels of additive white Gaussian noise and reconstructed with both AUTOMAP and conventional approaches: **a-c,** Radon projection encoding with SNR 40dB, compared with Kaczmarz-iterative Algebraic Reconstruction Technique (ART); **d-f,** Spiral *k*-space encoding with SNR 25 dB, compared with Conjugate Gradient SENSE reconstruction with NUFFT regridding; **g-i,** Poisson-disc undersampled (40%) Cartesian *k*-space encoding with SNR 30 dB, compared with compressed sensing reconstruction using the wavelet sparsifying transform. **j-l,** misaligned Cartesian *k*-space, compared with conventional inverse Fast Fourier Transform. In all cases, AUTOMAP demonstrates marked reduction of noise and reconstruction artifacts compared with conventional techniques.

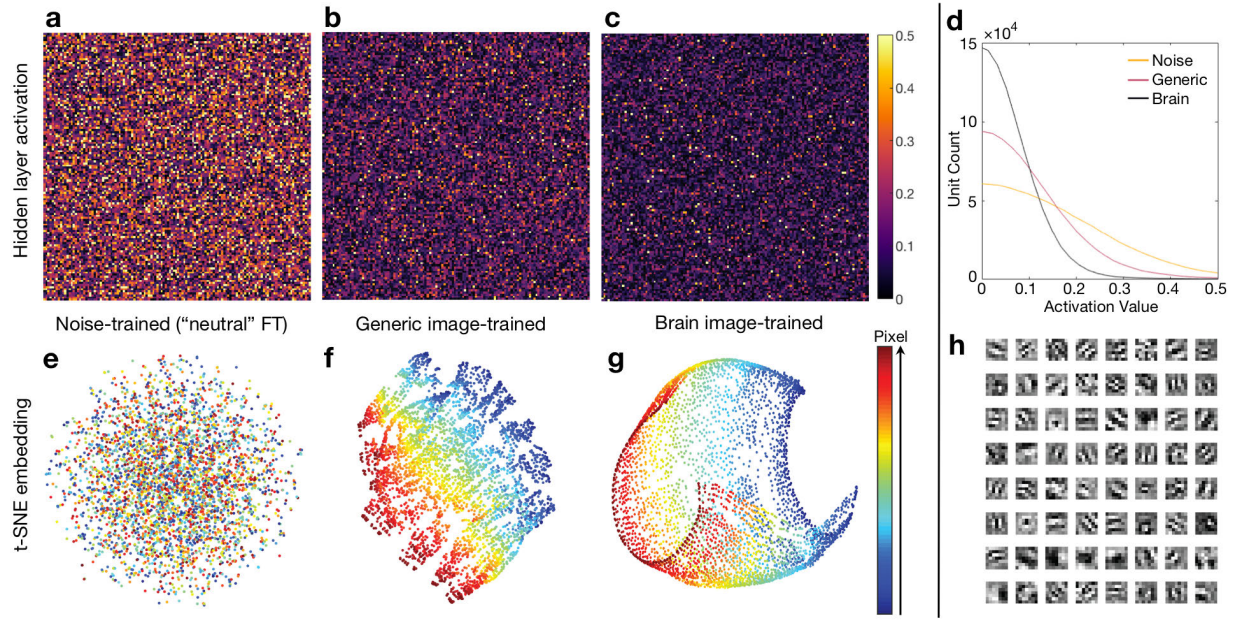

**Figure 3. Analysis of AUTOMAP neural networks.** AUTOMAP was trained on three separate datasets for a Cartesian *k*-space encoding: generic natural images, brain images, and random-valued noise without any real-world image structure (see Supplementary Methods for training details). **a-c,** Activation values of fully-connected hidden layer (FC2 in Fig. 1c) for each trained network while reconstructing the same *k*-space of a brain image. The noise-trained network generates high-amplitude and widely-distributed hidden layer activation values (a), while the networks trained on generic images (**b**) and brain images (**c**) exhibit greater sparsity, indicating efficient processing of input data due to successful feature extraction when trained on relevant data. **d,** Histogram of FC2 activation values for the three networks, accumulated over 100 brain image *k*-space reconstructions. **e-g,** Three-dimensional t-SNE embedding of network weights from FC2 to FC3 for the differently-trained networks (see Supplementary Methods for t-SNE analysis details). The t-SNE of the noise-trained network, agnostic to real-world image structure, exhibits disorganized structure (**e**), in contrast to (**f**) and (**g**), which reflect the local spatial correlation that exists in real-world images. The domain-specific training of the brain-trained network show the highest similarity between weights to two-dimensional neighbors for all pixel locations (**g**). **h,** The sparse convolutional kernels of the final convolutional stage (C2-Image) learned from training on brain images.

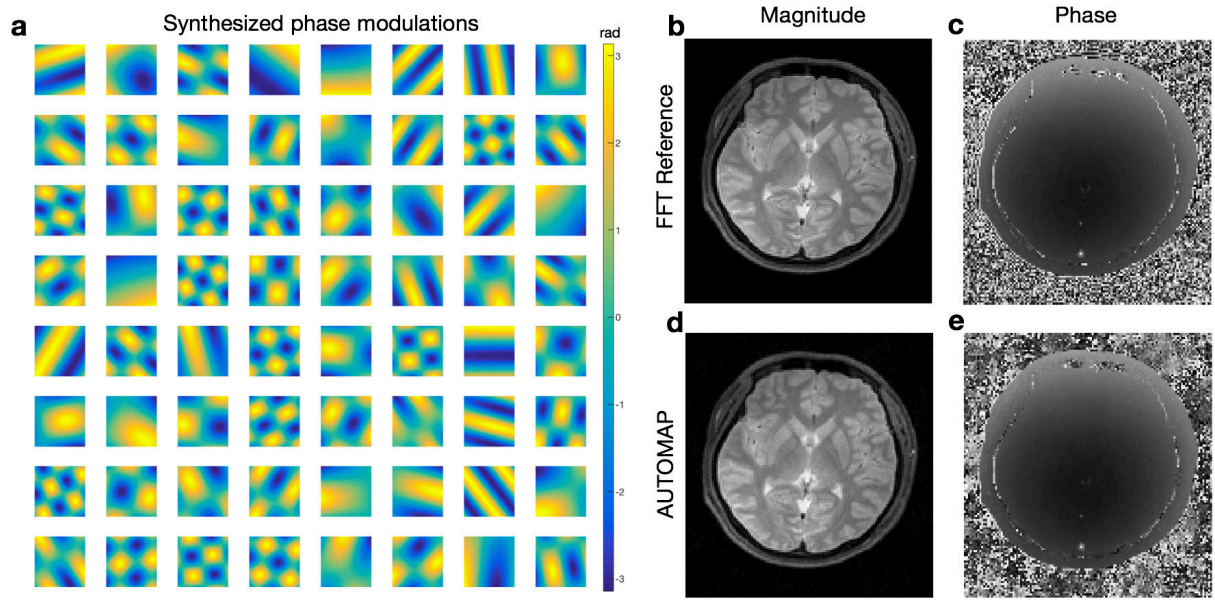

**Figure 4. Learning reconstruction of phase for *in vivo* data.** The inclusion of synthetic phase to the training dataset enables AUTOMAP to properly reconstruct both the magnitude and phase. **a,** The magnitude-only Human Connectome Project (HCP) *k*-space data was phase-modulated by two-dimensional sinusoids of varying spatial frequencies to generate the training dataset. After training, the magnitude (**b**) and phase (**c**) of a test T2-weighted *k*-space dataset are properly reconstructed by AUTOMAP (**d, e**).